\documentclass[runningheads]{llncs}

\usepackage{graphicx}
\usepackage[dvipsnames]{xcolor}

\usepackage[textsize=footnotesize,
disable
]{todonotes}
\usepackage[binary-units=true]{siunitx}

\usepackage{url}

\usepackage{subfig}

\usepackage[
hidelinks,
final % forces links, even if document is in draft
]{hyperref}
\hypersetup{
 pdftitle={ROS 2 for RoboCup},
 pdfsubject={There has always been much motivation for sharing code and solutions
  among teams in the RoboCup community. Yet the transfer of code
  between teams was usually complicated due to a huge variety of
  used frameworks and their differences in processing sensory
  information. The RoboCup@Home league has tackled this by transitioning
  to ROS as a common framework. In contrast, other leagues, such as those
  using humanoid robots, are reluctant to use ROS, as in those leagues real-time
  processing and low-computational complexity is crucial. However,
  ROS 2 now offers built-in support for real-time processing and
  promises to be suitable for embedded systems and multi-robot
  systems. It also offers the possibility to compose a set of nodes
  needed to run a robot into a single process. This, as we will show,
  reduces communication overhead and allows to have one single binary,
  which is pertinent to competitions such as the 3D-Simulation
  League. Although ROS 2 has not yet been announced to be production
  ready, we started the process to develop ROS 2 packages for using it
  with humanoid robots (real and simulated). This paper presents the
  developed modules, our contributions to ROS 2 core and RoboCup
  related packages, and most importantly it provides benchmarks that
  indicate that ROS 2 is a promising candidate for a common framework
  used among leagues.}, pdfauthor={Marcus M. Scheunemann, Sander
   G. van Dijk}, pdfkeywords={ROS 2, robot framework, robot software, embedded system, real-time system, minimal hardware, open source, humanoid robots, autonomous robots.} }

\begin{document}
\title{ROS 2 for RoboCup}
\author{Marcus M. Scheunemann \and Sander G. van Dijk}
\authorrunning{MM. Scheunemann and SG. van Dijk}
% First names are abbreviated in the running head.
% If there are more than two authors, 'et al.' is used.
%
\institute{University of Hertfordshire, AL10 9AB, UK\\
  \email{marcus@mms.ai} and \email{sgvandijk@gmail.com}\\
  \url{https://robocup.herts.ac.uk}}

\maketitle              % typeset the header of the contribution
\begin{abstract}
  There has always been much motivation for sharing code and solutions
  among teams in the RoboCup community. Yet the transfer of code
  between teams was usually complicated due to a huge variety of
  used frameworks and their differences in processing sensory
  information. The RoboCup@Home league has tackled this by transitioning
  to ROS as a common framework. In contrast, other leagues, such as those
  using humanoid robots, are reluctant to use ROS, as in those leagues real-time
  processing and low-computational complexity is crucial. However,
  ROS~2 now offers built-in support for real-time processing and
  promises to be suitable for embedded systems and multi-robot
  systems. It also offers the possibility to compose a set of nodes
  needed to run a robot into a single process. This, as we will show,
  reduces communication overhead and allows to have one single binary,
  which is pertinent to competitions such as the 3D-Simulation
  League. Although ROS~2 has not yet been announced to be production
  ready, we started the process to develop ROS~2 packages for using it
  with humanoid robots (real and simulated). This paper presents the
  developed modules, our contributions to ROS~2 core and RoboCup
  related packages, and most importantly it provides benchmarks that
  indicate that ROS~2 is a promising candidate for a common framework
  used among leagues. 
\end{abstract}
\section{Introduction}
Having a common framework among teams (or even among leagues) has many advantages. Most
notably, rather than concentrating on increasing performance or
reliability of the framework, participants can focus on the
implementation of artificial intelligence. Solutions can be easily
shared and distributed, with possibilities of benchmarking them
against each other. The handover and knowledge transfer within a team
to a new generation can be done smoothly. If there is a common
framework not just within a league but between leagues, then this may
foster the collaboration between teams of different leagues. With an
eye on the goal of merging forces of different leagues and eventually
merge leagues, a common framework is essential.

In a field such as a RoboCup league, where tasks and constraints
are similar for each participating team, one might think that a common
framework would naturally emerge. The RoboCup@Home Open Platform League
is a good example that this indeed happens. Starting with two teams in 2010,
in 2018 all teams announced within their team descriptions the use
of the same software framework~\cite{MatamorosSeibEtAl-18}.
The used framework was the open-source Robot Operating System~(ROS).
In other leagues, such as Standard Platform League~(SPL), ROS didn't
establish as a common framework. Performance shortcomings on minimal
hardware doesn't make it suitable for soccer playing
humanoid robots~\cite{RoeferLaue-14}. Instead, the self-developed
framework from team B-Human was adopted by many teams. The performance strength is gained through a
tight coupling between used software tools (e.g. simulator) and
between modules~\cite{RoeferLaue-14}.
However, this is also a shortcoming, as it restricts the community using the framework
mostly to the football playing domain.
NUClear is an example of a framework originating from RoboCup
that is more loosely coupled, modular and applicable in different
robot projects~\cite{HoulistonFountainEtAl-16}.
This framework solves the
overhead of traditional message passing systems through specially
optimised paths similar to those we will benchmark here. Additionally,
it offers the ability of using more blackboard/whiteboard type data
access patterns which are not directly available in the framework
discussed in this paper.

The authors themselves developed several frameworks with teams in
the Standard Platform League, Humanoid Kid-Size League and
3D~Simulation League. Their current
team, the Bold Hearts, used a self developed software framework, with
almost all modules created from
scratch~\cite{tdp-18}.
Although shown to be capable of performing well, over the years the
framework has become more and more complex. It is completely
custom and some of the original developers have moved on, making it
difficult to get new members started and to adapt it to new developments in the competitions. Code dependencies made it challenging to integrate well
working modules from other teams or projects. For example, last year's
change of the underlying vision pipeline towards a new semantic 
segmentation couldn't be easily achieved as third-party tools didn't
integrate well together~\cite{DijkScheunemann-18}.

There are reasons why some leagues couldn't agree on a common
framework yet. Naturally, without a central committee being in charge,
there will always be some healthy argument or a fork of a different,
perhaps slightly more efficient implementation. 

The fact that ROS is a framework from `outside' of RoboCup could
provide a good common base, but has
not been able to gain widespread adoption due to inherent limitations;
ROS was never built with support for, e.g., multi-robot systems
involving unreliable networks, for robots needing real-time processing
capabilities or for robots with minimal
hardware~\cite{FernandezFooteEtAl-14,RoeferLaue-14}.

However, exactly these limitations sparked the development of a
second, completely rewritten version of ROS. Although it is not yet
deemed fully production ready, it has a wider support than a RoboCup
team, a league or the whole community can offer: the support includes
large entities from the industry, such as Intel and Amazon. With an
eye on the future and the transferability of skills learned by our
(student) members outside/beyond participation in RoboCup, we opted to
use ROS~2~\cite{tdp-18}. Some benchmarking suggests that ROS~2 is
currently in a state that offers the possibility to use it for
multi-robot teams, small platforms and real-time
systems~\cite{GutierrezJuanEtAl-18}\todo{also mention as related 
work?}.

In this paper, we discuss why we think ROS~2 is a reasonable framework
choice and also briefly present its advantages~(Sec.~\ref{architecture}). We further present the modules
we have developed as a basis for participating in the
RoboCup~(Sec.~\ref{contributions}). Additionally, we present two
preliminary studies for benchmarking the system and showing the
feasibility for using it in RoboCup~(Sec.~\ref{benchmark}).

\section{ROS 2 Architecture and Features}
\label{architecture}
ROS~2 is based on the Data Distribution Service~(DDS) standard for
real-time systems~\cite{FernandezFooteEtAl-14}. DDS is a connectivity
framework aiming to enable scalability and real-time data exchange
using a Data-Centric Publish-Subscribe (DCPS)
architecture~\cite{OMG-15}.

DDS is specified by the Object Management Group~(OMG), which is an
open membership, not-for-profit computer industry standards
consortium. It is developed for a wide variety of fields such as
transportation systems, autonomous vehicles, and aerospace. ROS~2 sits
on top of that, providing standard messages and tools to adapt DDS for
robotic needs. A range of vendors providing implementations of DDS,
such as eProsima's Fast RTPS and RTI Connext are fully
supported~\cite{ROS-19}.
Compared to ROS~1, ROS~2 has several beneficial features:
\begin{itemize}
\item Built-in support for real-time systems.
\item Support for defining the `Quality of Service' of topics. This
  allows one to make a range of trade-offs between strong reliability
  and `best effort' policies, to deal with lossy communication. For
  instance, an efficient non-blocking `best effort', `UDP-like'
  service is acceptable for high frequency sensor data where missing
  individual messages is not detrimental. On the other hand, when each
  message is crucial, a reliable `TCP-like' policy can be used.
\item Nodes can be run in individual executables, or composed, using a
  variety of executors. In ROS~1, one has to maintain `nodelet'
  versions of all nodes to make this possible. In ROS~2, this can be
  achieved natively, making it possible to remove much of the
  communication overhead between nodes by having them share memory. We
  provide a benchmark showing the appeal of this below.
\item No need to run the ROS~1 \texttt{roscore} instance and maintain
  environment variables to make it and nodes reachable; with DDS,
  nodes discover each other through a network automatically.
\item Communication between nodes can be strictly restricted by
  placing them in different `domains'. This could be very useful in
  the RoboCup Simulation league for instance, where the programs of
  all robots in the same team run on the same machine but no
  communication between them should happen. Together with the ability
  to compose all modules for one agent in a single executable, this
  makes ROS~2 a much better candidate to use as a platform in the
  simulation league than ROS~1.
\end{itemize}

\section{ROS 2 and RoboCup Contributions}%
\label{contributions}%
\begin{figure}[htb]
  \centering
  \includegraphics[width=.80\textwidth]{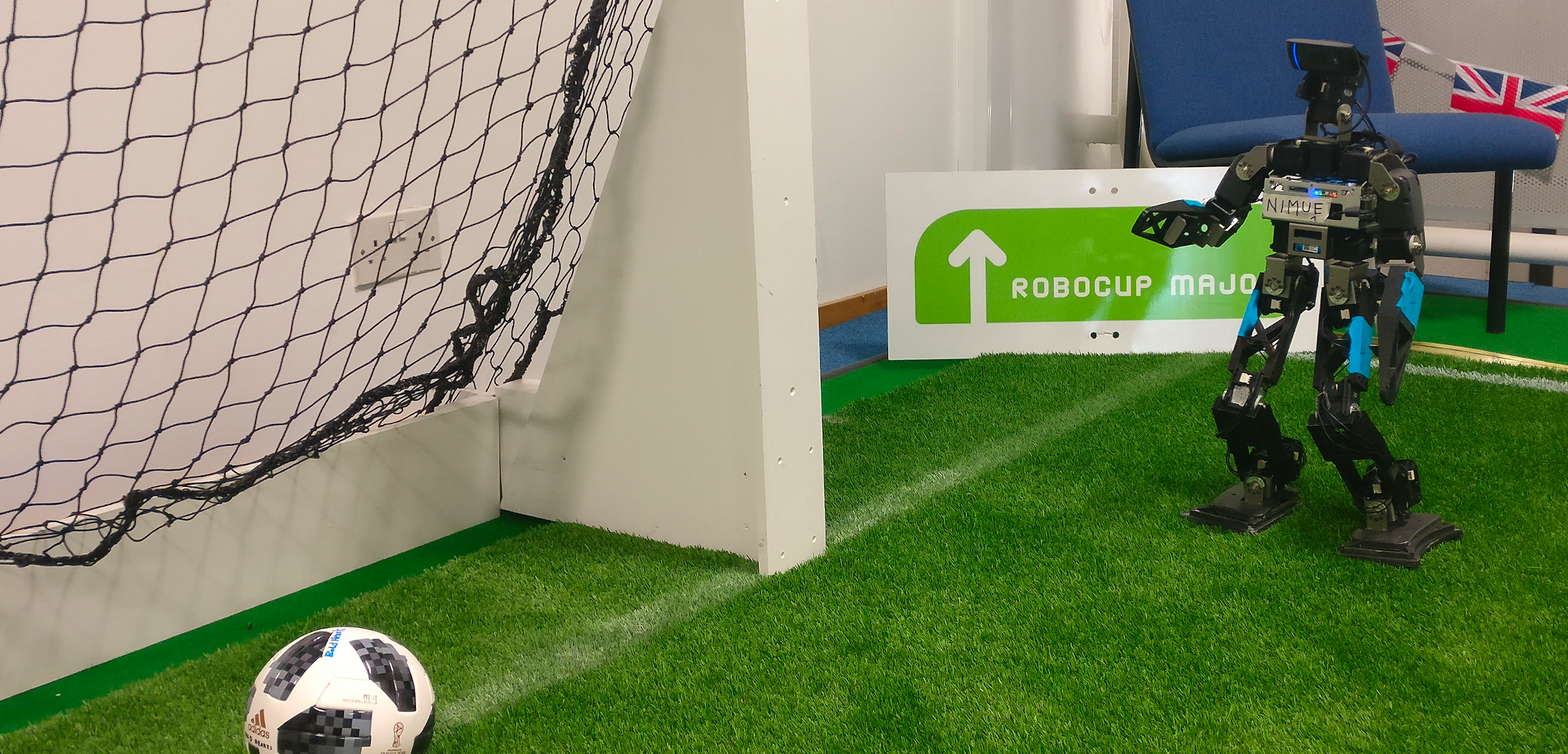}
  \includegraphics[width=.80\textwidth]{img/robot_camera_imu_rviz2_cut.png}
  \caption{The upper image depicts a scene with a robot looking at a
    ball. The lower image is a screenshot of RViz2. It shows
    the camera image feed retrieved with our USB camera driver~(left).
    Our CM-730 package publishes joint states, accelerometer and gyro information for
    building the robot model and compute its orientation with our IMU fusion package~(both right).
    }
  \label{fig:rviz2}
\end{figure}%
\begin{figure}[htb]%
  \centering
  \includegraphics[width=\textwidth]{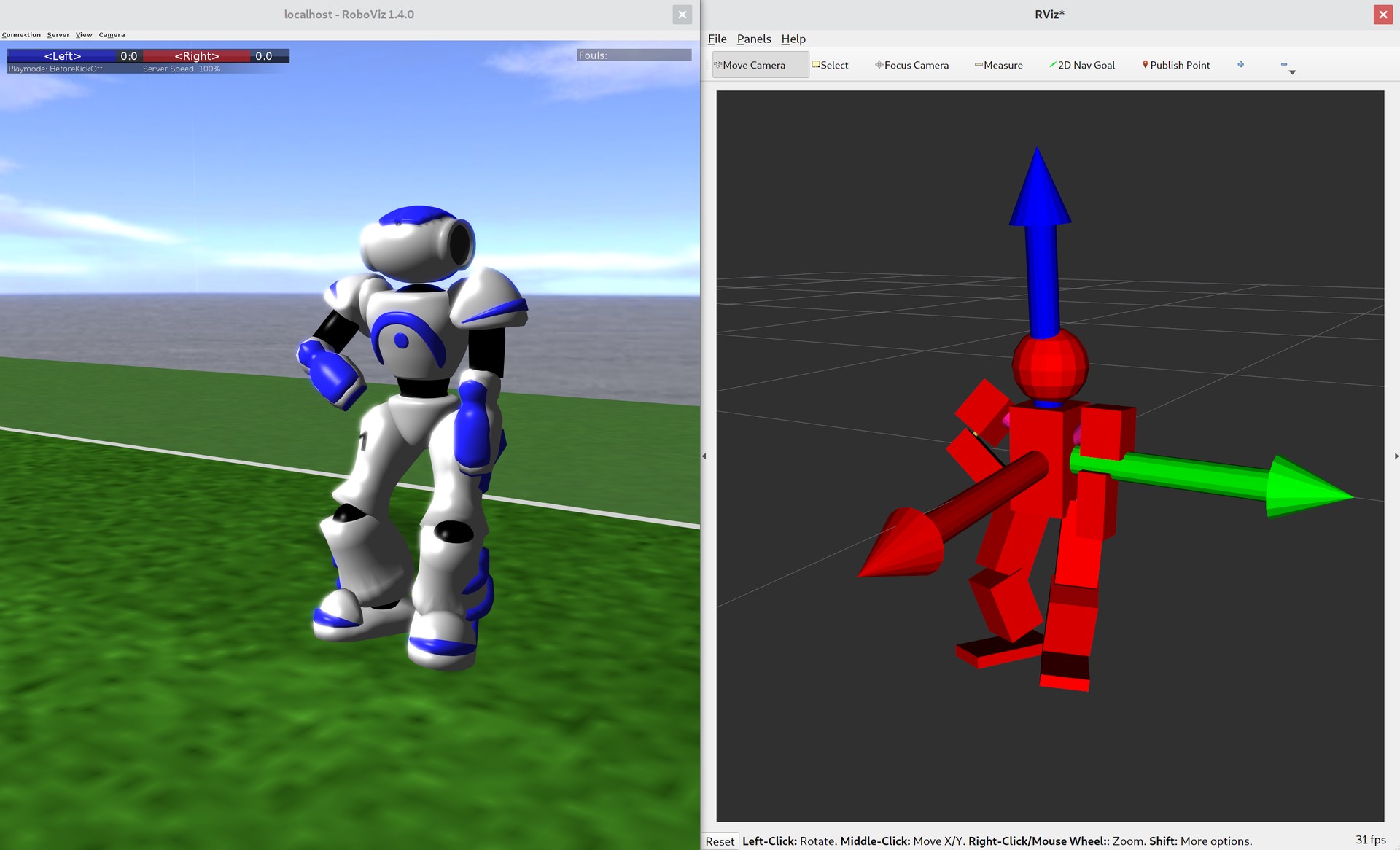}
  \caption{Depicted is a scene from RCSSServer3D used in the
    3D~Simulation League~(left, using RoboViz). Our package
    translates the servo information into standard ROS messages and
    publishes the topic \texttt{/joint\_state}. Also, the simulated
    gyroscope and accelerometer information are published.
    % TODO size restrictions
    Our IMU fusion package subscribes to the messages and computes the robot's
    orientation, exactly as for the real robot in Fig.~\ref{fig:rviz2}.
    The interface package allows for using ROS~2
    within the context of the 3D~Simulation League.}
  \label{fig:rcss3d}%
\end{figure}%
\noindent There are many modules that are not available yet, as ROS~2 is still relatively new. To be able to develop a full RoboCup team
based on ROS~2, we have developed several modules, consisting of:

\begin{description}
\item[Hardware driver] Our robots are based on the Robotis CM-730
  sub-controller. Robotis has released ROS~1 packages for their
  products, but at the moment there is no ROS~2 effort. We have
  created and published a ROS~2 driver for interacting with the
  CM-730, and controlling Robotis Dynamixel motors attached to
  it\footnote{\url{https://gitlab.com/boldhearts/ros2_cm730}}.
  Figure~\ref{fig:rviz2}~(right) shows the result of the robot model
  built using the output of this driver.
  We also developed a USB camera
  driver\footnote{\url{https://gitlab.com/boldhearts/ros2_v4l2_camera}},
  Figure~\ref{fig:rviz2}~(left) shows the output of the camera driver.
\item[Ports of our modules] With all hardware interfaces in place, we
  now work on porting our existing modules over to the new
  platform. The IMU fusion
  filter\footnote{\url{https://gitlab.com/boldhearts/ros2_imu_tools}}
  is one example. We complemented it with a package for visualizing the orientation
  of the robot in RViz2. Figure~\ref{fig:rviz2} and Figure~\ref{fig:rcss3d}~(both right)
  show the visualization of the orientation.
\item[Humanoid League] The RoboCup humanoid league uses a Game
  Controller application to manage a competition: it keeps track of
  and broadcasts the game state and events such as kick-off and
  penalties to the robots. We have created a package that forms a
  bridge between the communication protocol of the Game Controller and
  ROS~2
  topics\footnote{\url{https://gitlab.com/boldhearts/ros2_game_controller}}.
\item[3D Simulation League] Taking advantage of the benefits described
  above, we have developed a ROS~2 interface for the RoboCup
  3D~Simulation Server~\cite{XuVatankhah-14}\footnote{\url{https://gitlab.com/boldhearts/ros2_rcss3d}}.
  It uses the same platform and standard message interfaces as for our
  humanoid robots, making it easier for (new) members to experiment
  and improve our modules, and deploy them to real robots directly.
\item[ROS~2 core contributions] We have made several contributions to
  the core ROS~2 project for issues discovered in our use cases,
  including fixes to make it possible to compile ROS~2 for \SI{32}{\bit} ARM
  platforms\footnote{\url{https://github.com/ros2/rcl/pull/365}}, support modern Linux \SI{64}{\bit} library paths\footnote{\url{https://github.com/colcon/colcon-library-path/pull/10}} and to set
  complex node parameters using command line tools\footnote{\url{https://github.com/ros2/ros2cli/pull/199}}, along with smaller fixes for \texttt{geometry2}\footnote{\url{https://github.com/ros2/geometry2/pull/102}} and demo \texttt{image\_tools}\footnote{\url{https://github.com/ros2/demos/pull/288}}.
\end{description}

\section{Benchmark Stand-alone versus Composed Nodes}
\label{benchmark}
We use the humanoid football robot BoldBot for the benchmarks. Its
main board is an Odroid-XU4. This device is based on a Samsung
Exynos~5422 Cortex-A15 with \SI{2}{\giga\hertz} and a Cortex-A7 Octa
core CPU, which is the same as used in some 2015 model flagship
smartphones~\cite{tdp-18}. The main board runs \SI{32}{\bit}
Ubuntu~18.04 with a compiled version of ROS~2 Dashing Diademata, the
first long-term support version of ROS~2. All packages used in these
benchmarks and the ROS~2 core install have been compiled with GCC's
highest optimization level.

In ROS~1, nodes are stand-alone executables. Communication between
these nodes is performed through a transport protocol, most often over
TCP/IP. This means that all communication between these nodes involves
overhead from serialisation and memory copies. So called `nodelets'
were introduced to allow composing node-like building blocks into
single executables to alleviate such overhead. As these concepts are
separate from normal nodes, a package developer has to choose to
support either nodes or nodelets, or maintain both.

In ROS~2, nodes were redesigned to make it possible to run them either
stand-alone or composed in a single process, either single or
multi-threaded. ROS~2 Dashing Diademata even adds the ability to
dynamically load and unload nodes in a single process at runtime, as
so called `components'.

When nodes are composed in such a way, messages between them can be
shared directly, without any intermediate conversion overhead. ROS~2
explicitly supports this form of Intra-Process Communication~(IPC), 
bypassing the DDS layer and performing zero-copy
communication where possible. To test the impact of this, we perform
several benchmarks of one of the most memory intensive operations for
robots: image processing.

\subsection{Case 1: Simple Node Graph}%
\begin{figure}[htb]%
  \centering
  \includegraphics[width=\textwidth]{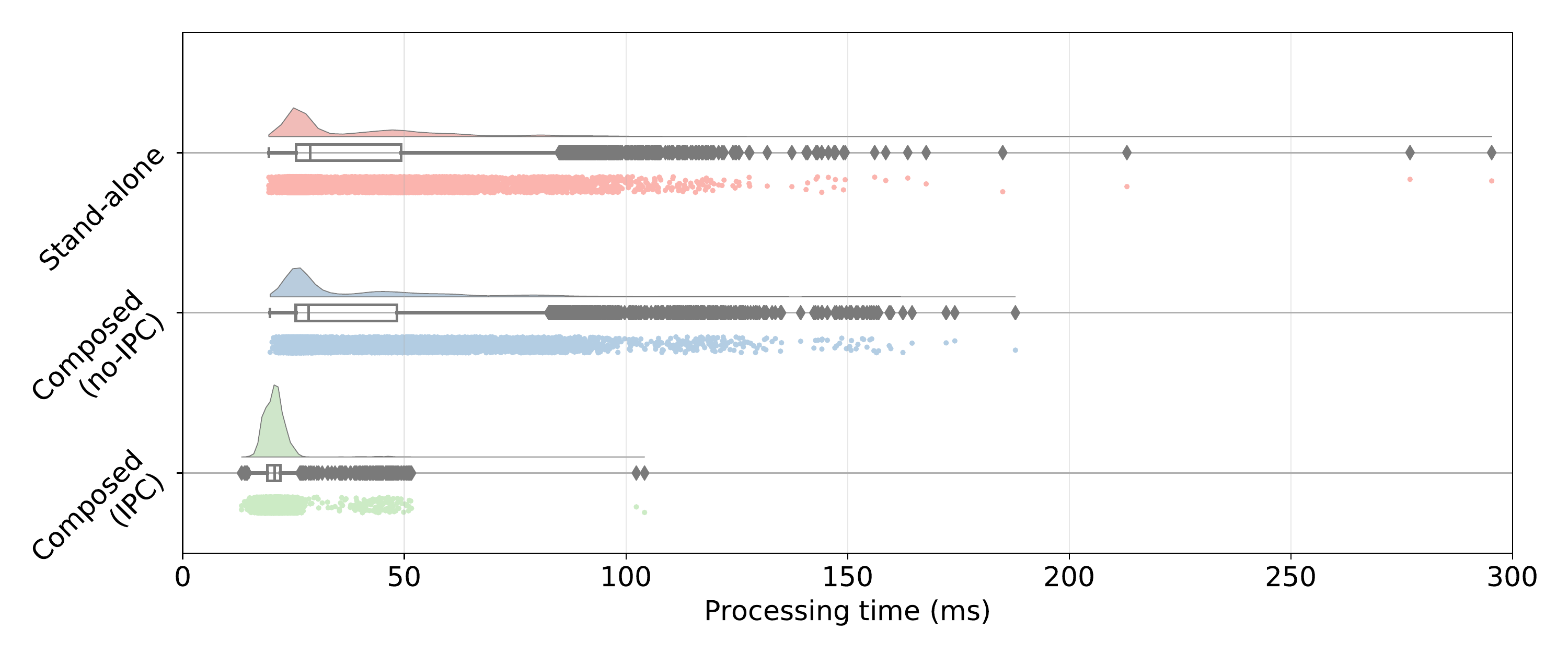}\label{fig:standalone-vs-composed-release}
  \caption{Distributions of time measured from image capture by camera
    node until end of processing of a Sobel gradient operation on the
    full image done by processing node. Both
    nodes either run as stand-alone executables or run composed in a
    single executable, the latter with IPC disabled and enabled. Each plot
    shows the density~(top), a boxplot~(middle), and individual data
    points~(bottom). 10.000~samples are measured in each
    case.}\label{fig:standalone-vs-composed}
\end{figure}%

\noindent A node is created that subscribes to the \texttt{/image\_raw} topic
provided by the camera driver described earlier. This node applies a
Sobel gradient operation to incur some actual processing cost to
offset overhead costs against, and measures the time from initial
image capture (provided in the \texttt{Image} header) until it
finished processing the image.

Figure~\ref{fig:standalone-vs-composed} shows the effects of composing
nodes and intra-process communication. The end-to-end processing time
for 10.000~images was compared when running the camera and processing
nodes separately, or when composed in a single executable using a
multi-threaded executor\footnote{When using a single-threaded executor
  no processing actually happened, possibly due to the camera node
  claiming all execution time}, the latter with IPC disabled and
enabled. Composing the nodes seems to result in slightly more stable
communication, with less extreme outliers than in the stand-alone
case, but the difference is minimal.

However, a clear benefit can be seen when IPC is enabled, with the
median processing time dropping by 33\% compared to the stand-alone
results, from \SI{28.7}{\milli\second} to
\SI{20.7}{\milli\second}. The communication is also much more
reliable, with a distinctly narrower distribution of processing
times. The zero-copy transmission that is responsible for this
improvement is achieved in ROS~2 by passing direct memory pointers to
messages from publisher to subscriber, when using the C++ `unique
pointer' concept to signal that this is safe to do.

\subsection{Case 2: Extended Node Graph}%
\begin{figure}[htb]
  \centering
  \includegraphics[width=\textwidth]{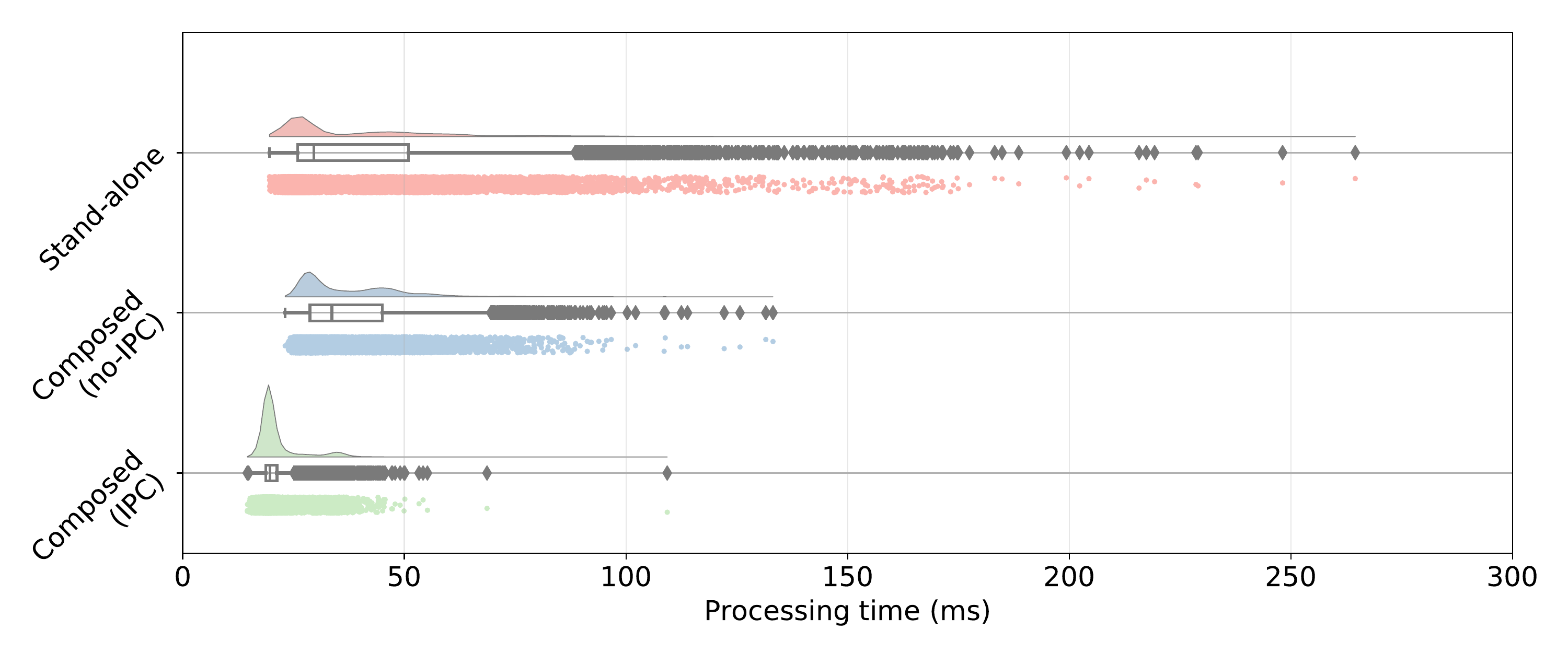}
  \caption{Image processing times as in
    Fig.~\ref{fig:standalone-vs-composed}, but with the robot system
    extended with other necessary nodes and topics like
    \texttt{/joint\_states} and \texttt{/imu/data}, for a total of 7
    nodes running at a time. Each plot shows the density~(top), a
    boxplot~(middle), and individual data points~(bottom).  10.000
    samples are measured in each
    case.}\label{fig:standalone-vs-composed_with_motion}
\end{figure}%

\noindent We extend the previous benchmark example to further
understand the impact of communication overhead in a more realistic
configuration with a larger set of nodes. To extend the image
processing example, we further introduce nodes to publish and process
IMU sensor and joint data, provided by our packages described in
Sec.~\ref{contributions}. Our CM-730 package reads the servo
information from the robot and publishes it as a standard topic
\texttt{/joint\_state}. It also publishes raw IMU readings of the
accelerometer and the gyroscope. The IMU fusion package reads these
raw messages and computes the robots' orientation and publishes the
\texttt{/imu/data} topic. Altogether, 7 nodes are involved that
communicate on 7 topics (including the image topic). For the benchmark
using composition and IPC, only the \texttt{/image\_raw} topic is
IPC-enabled.

\noindent Figure~\ref{fig:standalone-vs-composed_with_motion} shows the
results. The median computation time for the image processing for the
stand-alone binary increases by \SI{0.8}{\milli\second}, only a slight
increase when compared to the simple case from the previous section.
However, the tail of the distribution has become longer, indicating
further increase in communication variability.

This same effect is not seen in the composed, no-IPC results; the
distribution is actually tighter, although the median processing time
has increased from \SI{28.4}{\milli\second} to
\SI{33.6}{\milli\second}. We do not know the reason for this change
compared to the simple system. However, given the results when IPC
\emph{is} enabled, this mode is not recommended anyway.

This is because also in this extended case, enabling IPC gives
significantly better results; hardly any effect of adding more nodes
is visible in the image processing times, with the median time
even \SI{1}{\milli\second} lower than in the simple case.
These results show that a composed binary can prevent much of possible
costs of creating a modular system without having to maintain any
additional code, in contrast to ROS~1 for instance.

\section{Conclusion \& Future work}
We believe a common framework will ease the process of sharing and
comparing solutions between teams and help them concentrate on their
research interests.
This is valuable both within and between leagues. For instance,
ultimately the simulation league aims more for strategic play, whereas
the humanoid leagues focus heavily on developing robot hardware and
lower level control. Given the goal of merging these leagues
eventually, a common framework is an inevitable basis.

In this paper, we propose ROS~2 as a suitable choice for a framework
for a RoboCup team with needs for real-time processing relying on
minimal hardware. ROS~2 is supported by a large community, including
big industrial partners, that the RoboCup community can benefit
from. This makes it a potential candidate for a common framework. We
presented our ROS~2 and RoboCup
contributions, allowing to start with ROS~2 in the RoboCup.
We supported the CM-730 sub-controller manufactured by Robotis,
a very popular brand among RoboCup teams in the Humanoid League.
Furthermore, we wrote an interface for using ROS~2 with the simulator
of the 3D-Simulation League.

Our benchmarks indicate that ROS~2's capability to compose nodes
can reduce the communicational overhead known from ROS~1. Our future
work mostly includes porting over ROS~1 packages and our custom modules
to the ROS~2 ecosystem.

%
% ---- Bibliography ----
%
% BibTeX users should specify bibliography style 'splncs04'.
% References will then be sorted and formatted in the correct style.
%

\end{document}